\documentclass[10pt,twocolumn,letterpaper]{article}

\usepackage{cvpr}
\usepackage{times}
\usepackage{epsfig}
\usepackage{graphicx}
\usepackage{amsmath}
\usepackage{amssymb}

\usepackage[breaklinks=true,bookmarks=false]{hyperref}
\usepackage[multiple]{footmisc}

\cvprfinalcopy 

\def\cvprPaperID{****} 

\begin{document}

\title{Face Attention Network: An Effective Face Detector for the Occluded Faces}


\author{
Jianfeng Wang\thanks{Equal contribution.} \thanks{Work was done during an internship at Megvii Research.}\\
College of Software, Beihang University\\
Beijing, China\\
{\tt\small wjfwzzc@buaa.edu.cn}
\and
Ye Yuan\footnotemark[1]\\
Megvii Inc. (Face++)\\
Beijing, China\\
{\tt\small yuanye@megvii.com}
\and
Gang Yu\\
Megvii Inc. (Face++)\\
Beijing, China\\
{\tt\small yugang@megvii.com}
}

\maketitle




\begin{abstract}
The performance of face detection has been largely improved with the development of convolutional neural network. However, the occlusion issue due to mask and sunglasses, is still a challenging problem. The improvement on the recall of these occluded cases usually brings the risk of high false positives. In this paper, we present a novel face detector called Face Attention Network (FAN), which can significantly improve the recall of the face detection problem in the occluded case without compromising the speed. More specifically, we propose a new anchor-level attention, which will highlight the features from the face region. Integrated with our anchor assign strategy and data augmentation techniques, we obtain state-of-art results on public face detection benchmarks like WiderFace and MAFA. The code will be released for reproduction.
\end{abstract}

\section{Introduction}

Face detection is a fundamental and essential step for many face related applications, e.g. face landmark~\cite{xiong2013supervised, zhu2016face} and face recognition~\cite{parkhi2015deep,schroff2015facenet,zhu2015high}. Starting from the pioneering work of Viola-Jones~\cite{viola2004robust}, face detection witnessed a large number of progress, especially as the recent development of convolutional neural networks~\cite{He2015}. However, the occlusion problem is still a challenging problem and few of the work have been presented to address this issue. More importantly, occlusion caused by mask, sunglasses or other faces widely exists in the real-life applications. 

The difficulty to address the occlusion issue lies at the risk of potential false positive problem. Considering the case of detecting a face occluded by a sunglasses, only the lower part of face is available. The models, which can recognize the face only based on the lower part, will be easily misclassified at the positions like hands which share the similar skin color. How to successfully address the occlusion issue and meanwhile prevent the false positive problem is still a challenging research topic.

In this paper, we present an effective face detector based on the one-shot detection pipeline called Face Attention Network (FAN), which can well address the occlusion and false positive issue. More specifically, following the similar setting as RetinaNet~\cite{lin2017focal}, we utilize feature pyramid network, and different layers from the network to solve the faces with different scales. Our anchor setting is designed specifically for the face application and an anchor-level attention is introduced which provides different attention regions for different feature layers. The attention is supervised trained based on the anchor-specific heatmaps. In addition, data augmentation like random crop is introduced to generate more cropped (occluded) training samples.

In summary, there are three contributions in our paper. 
\begin{itemize}
\item We propose a anchor-level attention, which can well address the occlusion issue in the face detection task. One illustrative example for our detection results in the crowd case can be found in Figure~\ref{fig:view}.
\item A practical baseline setting is introduced based on the one-shot RetinaNet detector, which obtains comparable performance with fast computation speed. 
\item Our FAN which integrates our reproduced one-shot RetinaNet and anchor-level attention significantly outperforms state-of-art detectors on the popular face detection benchmarks including WiderFace~\cite{yang2016wider} and MAFA~\cite{Ge_2017_CVPR}, especially in the occluded cases like MAFA.
\end{itemize}

\section{Related Work}\label{sec:relatedwork}
Face detection as the fundamental problem of computer vision, has been extensively studied. Prior to the renaissance of convolutional neural network (CNN), numerous of machine learning algorithms are applied to face detection. The pioneering work of Viola-Jones~\cite{viola2004robust} utilizes Adaboost with Haar-like feature to train a cascade model to detect face and get a real-time performance. Also, deformable part models (DPM)~\cite{felzenszwalb2008discriminatively} is employed for face detection with remarkable performance. However, A limitation of these methods is that their use of weak features, e.g., HOG~\cite{dalal2005histograms} or Haar-like features~\cite{viola2001rapid}. Recently, deep learning based algorithm is utilized to improve both the feature representation and classification performance. 
Based on the whether following the proposal and refine strategy, these methods can be divided into: single-stage detector, such as YOLO~\cite{redmon2016you}, SSD~\cite{liu2016ssd}, RetinaNet~\cite{lin2017focal}, and two-stage detector such as Faster R-CNN~\cite{ren2015faster}.

\textbf{Single-stage method.} CascadeCNN~\cite{li2015convolutional} proposes a cascade structure to detect face coarse to fine. MT-CNN~\cite{zhang2016joint} develops an architecture to address both the detection and landmark alignment jointly. Later, DenseBox~\cite{huang2015densebox} utilizes a unified end-to-end fully convolutional network to detect confidence and bounding box directly. UnitBox~\cite{yu2016unitbox} presents a new intersection-over-union (IoU) loss to directly optimize IOU target. SAFD~\cite{hao2017scale} and RSA unit~\cite{liu2017recurrent} focus on handling scale explicitly using CNN or RNN. RetinaNet~\cite{lin2017focal} introduces a new focal loss to relieve the class imbalance problem.  

\textbf{Two-stage method.} Beside, face detection has inherited some achievements from generic object detection tasks. \cite{jiang2017face} use the Faster R-CNN framework to improve the face detection performance. CMS-RCNN~\cite{zhu2017cms} enhances Faster R-CNN architecture by adding body context information. Convnet~\cite{li2016face} joins Faster R-CNN framework with 3D face model to increase occlusion robustness. Additionally, Spatial Transformer Networks (STN)~\cite{jaderberg2015spatial} and it's variant~\cite{chen2016supervised}, OHEM~\cite{shrivastava2016training}, grid loss~\cite{opitz2016grid} also presents several effective strategies to improve face detection performance.

\begin{figure*}[h]
\begin{center}
    \includegraphics[width=0.75\linewidth]{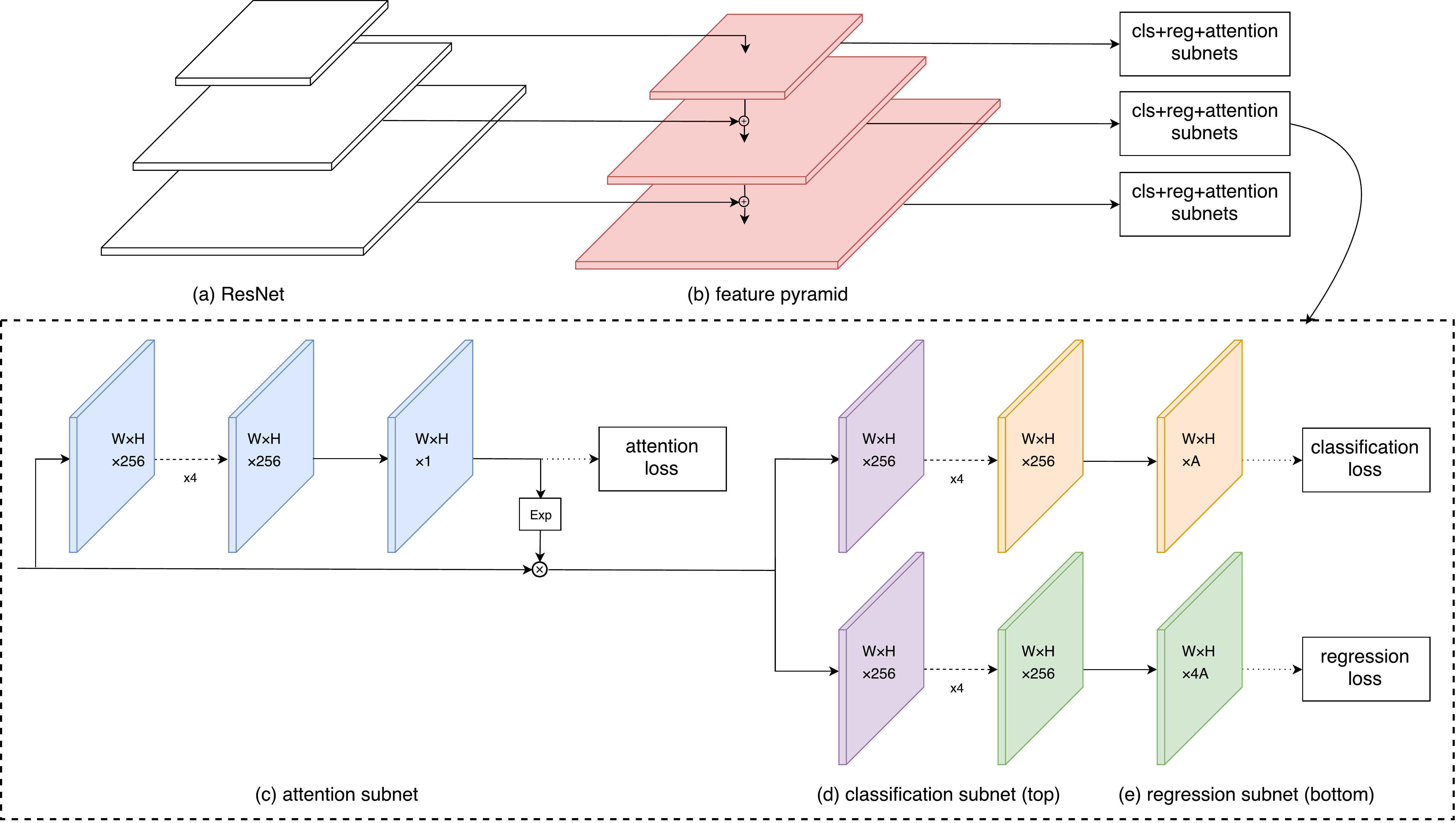}
\end{center}
    \caption{An overview of our FAN network architecture.}
\label{fig:architecture}
\end{figure*}

\section{Face Attention Network~(FAN)}
Although remarkable improvement have been achieved for the face detection problem as discussed in Section~\ref{sec:relatedwork}, the challenge of locating faces with large occlusion remains. Meanwhile, in many face-related applications such security surveillance, faces usually appear with large occlusion by mask, sunglasses or other faces, which need to be detected as well. \cite{najibi2017ssh} addresses this problem by merging information from different feature layers. \cite{zhang2017s3fd} changes anchor matching strategy in order to increase recall rate. Inspire of~\cite{dollar2014fast}, we are trying to leverage a fully convolutional feature hierarchy, and each layer targets to handle faces with different scale range by assigning different anchors. Our algorithm, called Face Attention Network (FAN), can be considered as an integration of a single-stage detector discussed in Section~\ref{sec:base setting} and our anchor-level attention in Section~\ref{sec:attention network}. An overview of our network structure can be found in Figure~\ref{fig:architecture}.

\subsection{Base Framework\label{sec:base setting}}

Convolutional neural network has different semantic information and spatial resolution at different feature layers. The shallow layers usually have high spatial resolution, which is good for spatial localization of small objects, but low semantic information, which is not good for visual classification. On the other hand, deep layers obtain more semantic information but the spatial resolution is compromised. Recent work like Feature Pyramid Network (FPN)~\cite{lin2016feature} proposes a divide and conquer principle. A U-shape structure is attached to maintain both the high spatial resolution and semantic information. Different scales of objects are split and addressed at different feature layers.

Following the design principle, RetinaNet introduced in~\cite{lin2017focal} is a one-stage detector which achieves state-of-art performance on COCO general object detection~\cite{lin2014microsoft}. It employs FPN~\cite{lin2016feature} with ResNet~\cite{He2015} as backbone to generate a hierarchy of feature pyramids with rich semantic information. Based on this backbone, RetinaNet is attached with two subnets: one for classifying and the other for regressing. We borrow the main network structure from RetinaNet and adapt it for the face detection task.
 
The classification subnet applies four $ 3\times3 $ convolution layers each with 256 filters, followed by a $ 3\times3 $ convolution layer with $ K A $ filters where $ K $ means the number of classes and $ A $ means the number of anchors per location. For face detection $ K=1 $ since we use sigmoid activation, and we use $ A=6 $ in most experiments. All convolution layers in this subnet share parameters across all pyramid levels just like the original RetinaNet. The regression subnet is identical to the classification subnet except that it terminates in $ 4A $ convolution filters with linear activation. Figure~\ref{fig:architecture} provides an overview for our algorithm. Notice, we only draw three levels pyramids for illustrative purpose.

\subsection{Attention Network}\label{sec:attention network}

Compared with the original RetinaNet, we have designed our anchor setting together with our attention function. There are three design principles:
\begin{itemize}
\item addressing different scales of the faces in different feature layers, 
\item highlighting the features from the face region and diminish the regions without face,
\item generating more occluded faces for training. 
\end{itemize}

\subsubsection{Anchor Assign Strategy\label{sec:anchor matching}}

\begin{figure}[h]
\begin{center}
    \includegraphics[width=0.8\linewidth]{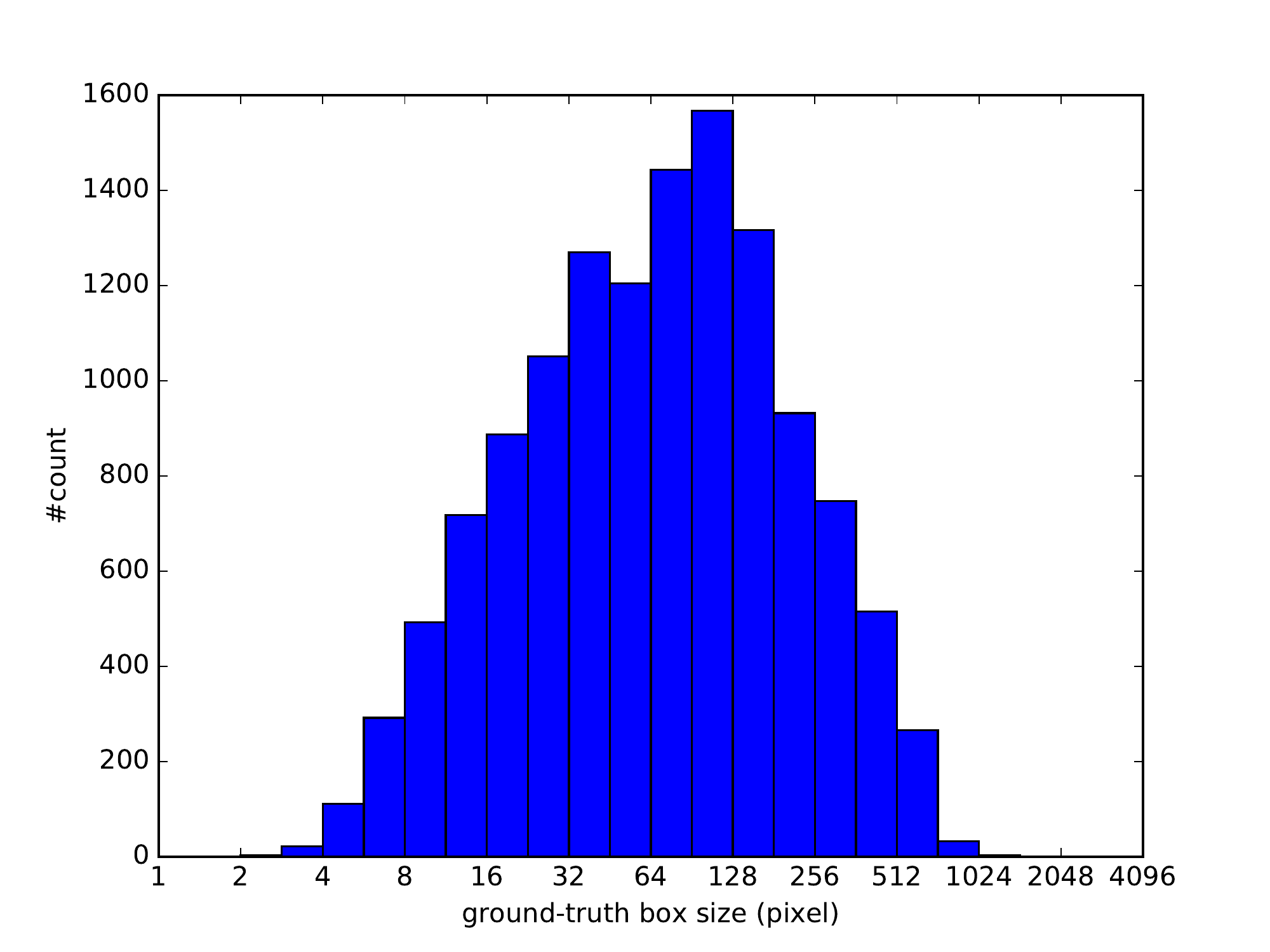}
\end{center}
    \caption{The distribution of object scales in the WiderFace train dataset More than 80\% categories have a object
size between 16 and 406 pixel.}
\label{fig:widerface}
\end{figure}

We start the discussion of our anchor setting first. In our FAN, we have five detector layers each associated with a specific scale anchor. In addition, the aspect ratio for our anchor is set as 1 and 1.5, because most of frontal faces are approximately square and profile faces can be considered as a 1:1.5 rectangle. Besides, we calculate the statistics from the WiderFace train set based on the ground-truth face size. As Figure~\ref{fig:widerface} shows, more than 80\% faces have an object scale from 16 to 406 pixel. Faces with small size lack sufficient resolution and therefore it may not be a good choice to include in the training data. Thus, we set our anchors from areas of $ 16^2 $ to $ 406^2 $ on pyramid levels. We set anchor scale step to $ 2 ^ {1/3} $, which ensure every ground-truth boxes have anchor with IoU $ \geq 0.6 $. Specifically, anchors are assigned to a ground-truth box with the highest IoU larger than $0.5$, and to background if the highest IoU is less than $0.4$. Unassigned anchors are ignored during training.

\subsubsection{Attention Function\label{sec:attention func}}

\begin{figure}[t]
\begin{center}
    \includegraphics[width=0.9\linewidth]{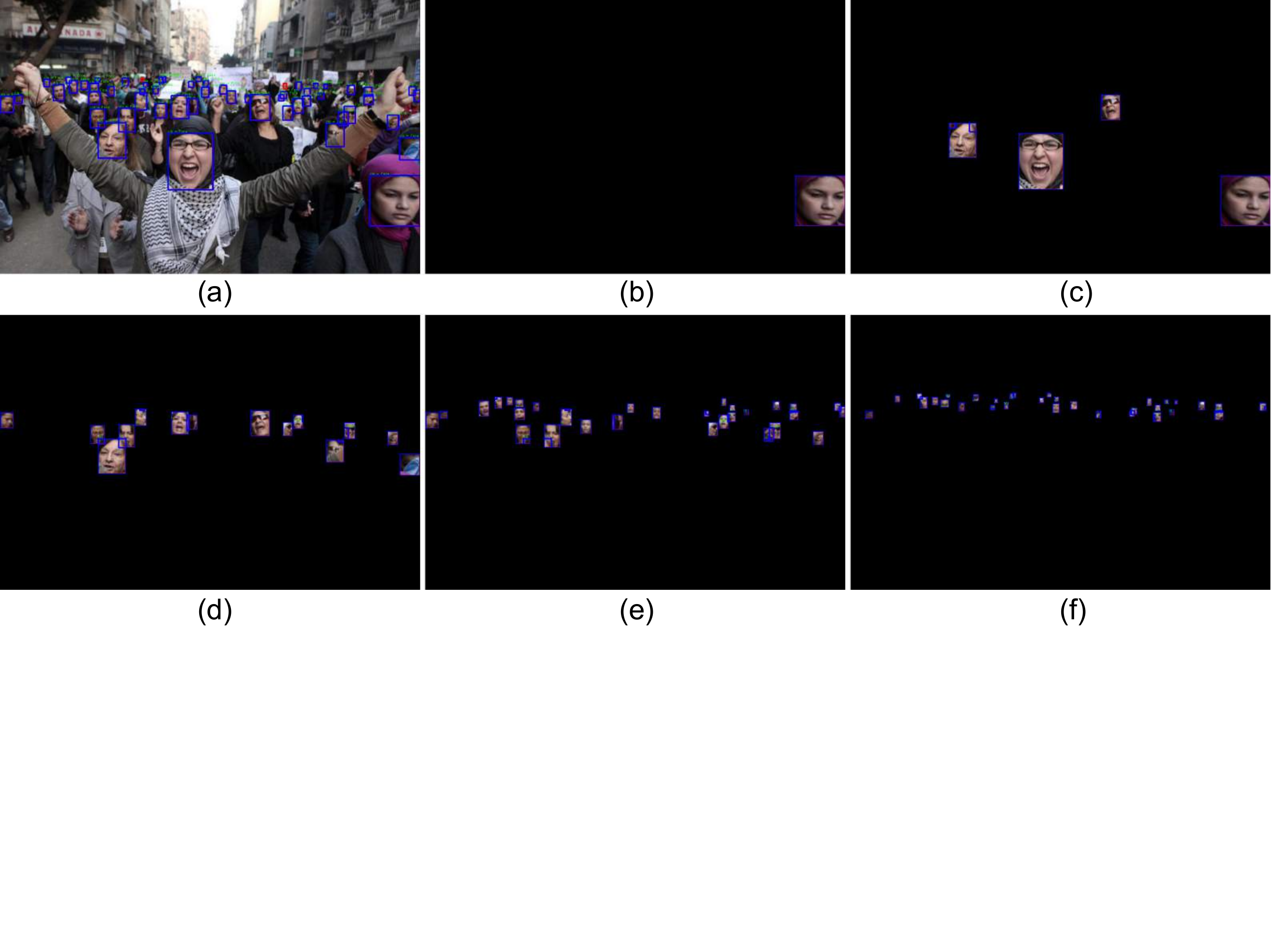}
\end{center}
    \vskip -2cm
    \caption{Attention supervised information is associated to the ground-truth which the anchors at the current layer match. Different layers have different ground-truth based on the face size. The subfigures refer to: (a) the origin image with ground-truth (b) the attention supervised information of P7, which care for large face, (c) the attention supervised information of P6 layer, it is smaller than P7. (d)-(f) the ground-truth assigned for P5 to P3, which focus on smaller face, respectively.}
\label{fig:attention_gt}
\end{figure}

\begin{figure}[t]
\begin{center}
    \includegraphics[width=0.9\linewidth]{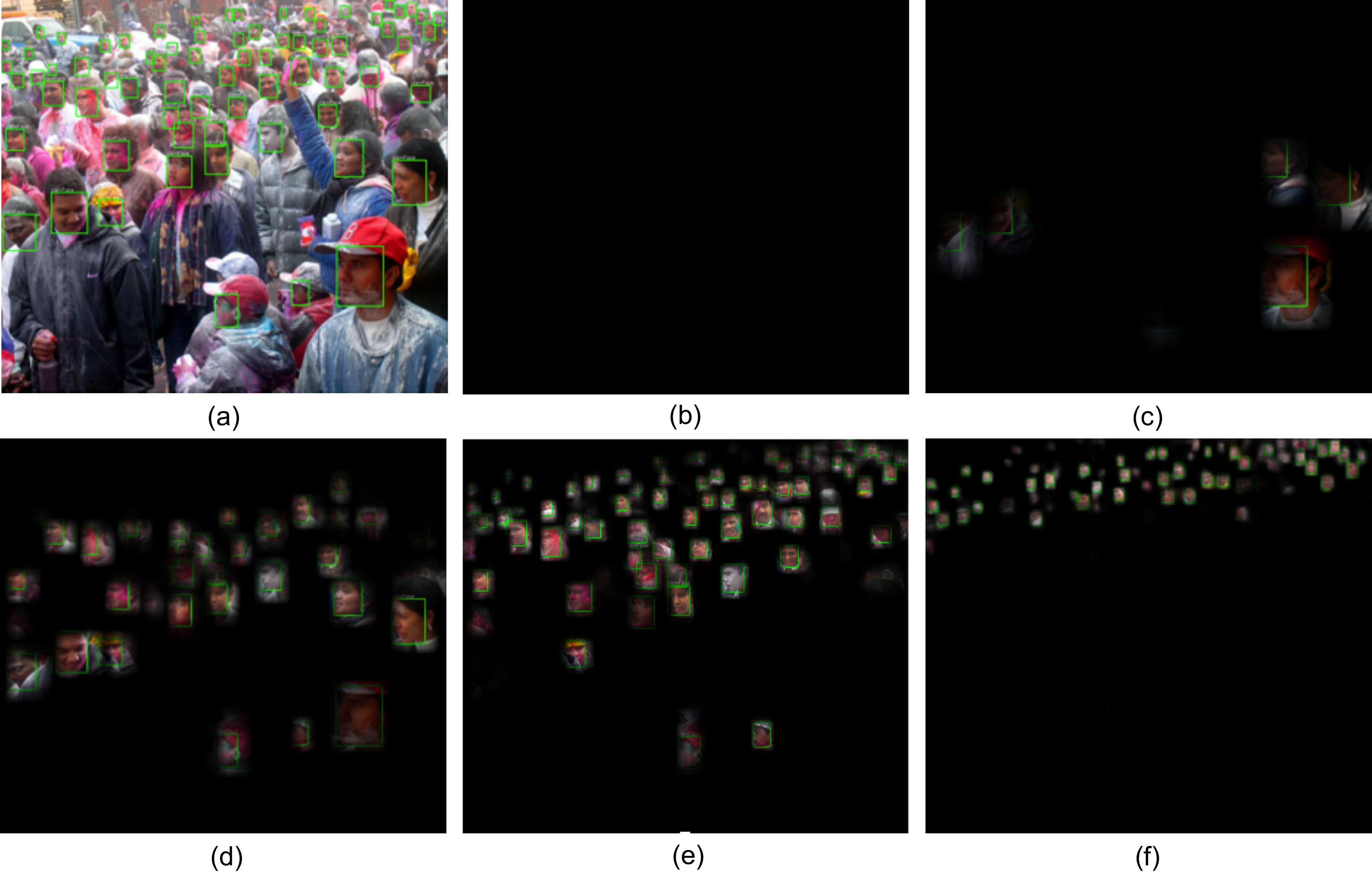}
\end{center}
    \caption{The original image (a) and (b)-(f) present attention maps for P7 to P3 produced by our FAN, respectively. Figures are from WiderFace validation dataset.} 
\label{fig:occlusion_attention}
\end{figure}

\begin{figure*}[h]
\begin{center}
    \includegraphics[width=0.9\linewidth]{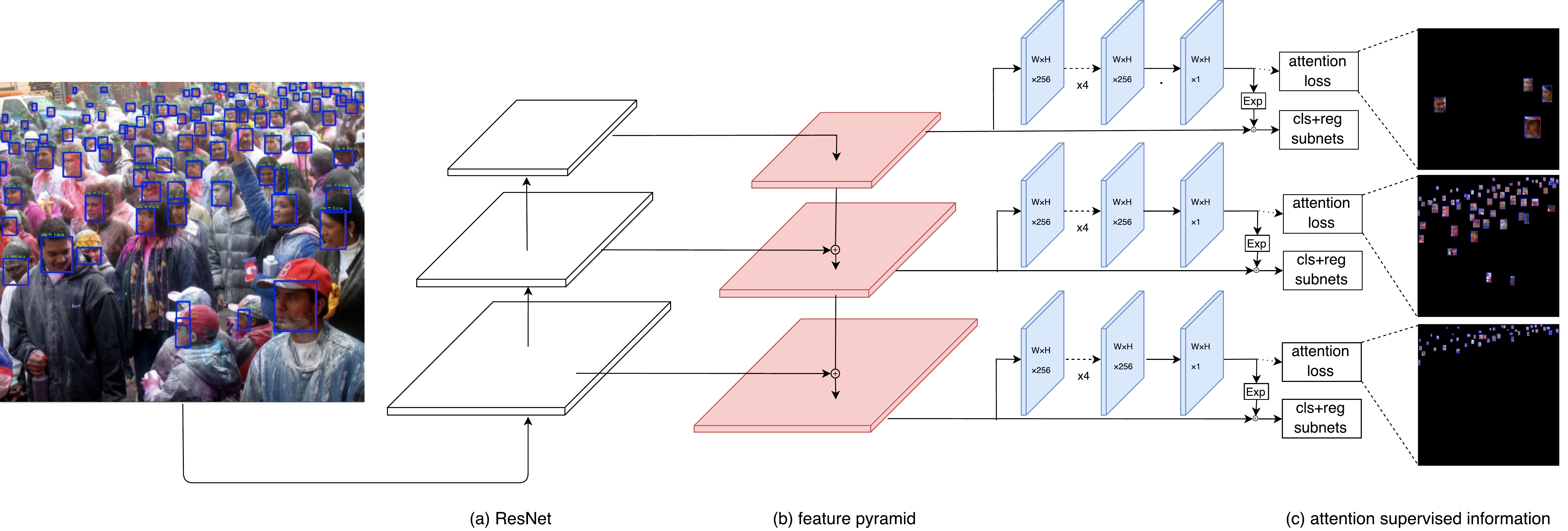}
\end{center}
    \caption{An overview of our hierarchy attention mechanism for training. Different from traditional usage of attention map, our attention maps are first fed to an exponential operation and then dot with feature maps. Besides, the attention supervision information is associated to the ground-truth faces assigned to the anchors in the current layer.}
\label{fig:hierarchy_attetion}
\end{figure*}

To address the occlusion issue, we propose a novel anchor-level attention based on the network structure mentioned above. Specifically, we utilize a segment-like side branch as shown in Figure~\ref{fig:hierarchy_attetion}. The attention supervision information is obtained by filling the ground-truth box. Meanwhile as Figure~\ref{fig:attention_gt} shows, supervised heatmaps are associated to the ground-truth faces assigned to the anchors in the current layer. These hierarchical attention maps could decrease the correlation among them. Different from traditional usage of attention map, which naively multiple it with the feature maps, our attention maps are first feed to an exponential operation and then dot with feature maps. It is able to keep more context information, and meanwhile highlight the detection information. Considering the example with occluded face, most invisible parts are not useful and may be harmful for detection. Some of the attention results can be found in Figure~\ref{fig:occlusion_attention}. Our attention mask can enhance the feature maps in the facial area, and diminish the opposition.

\subsubsection{Data Augmentation\label{sec:data_aug}}
We find that the number of occluded faces in the training dataset, e.g., WiderFace train, is limited and cannot satisfy the training of CNN network.
Only 16\% faces are with highly occlusion property from the annotation. Thus, we propose a random crop strategy, which can generate a large number of occluded faces for training. More specifically, based on the training set,  we randomly crop square patches from original images, whose range between [0.3, 1] of the short edge from the original images. In addition, We keep the overlapped part of the ground-truth box if its center is in the sampled patch. Besides from the random crop dataset augmentation, we also employ augmentation from random flip and color jitter~\cite{zhang2017s3fd}.

\subsection{Loss function}\label{sec:training}



We employ a multi-task loss function to jointly optimize model parameters:
\begin{equation}
\begin{aligned} L=&\sum_k{\frac{1}{N_k^c}\sum_{i\in A_k}{L_c(p_i, p_i^*)}}+\\
&\lambda_1\sum_k{\frac{1}{N_k^r}\sum_{i\in A_k}{I(p_i^*=1)L_r(t_i, t_i^*)}}+\\
&\lambda_2\sum_k{L_a(m_k, m_k^*)} \end{aligned},
\end{equation}
where k is the index of an feature pyramid level ($ k\in[3, 7] $), and $ A_k $ represents the set of anchors defined in pyramid level $ P_k $. The ground-truth label $ p_i^* $ is 1 if the anchor is positive, 0 otherwise. $ p_i$ is the predicted classification result from our model. $ t_i $ is a vector representing the 4 parameterized coordinates of the predicted bounding box, and $ t_i^* $ is that of the ground-truth box associated with a positive anchor. 

The classification loss $ L_c(p_i, p_i^*) $ is focal loss introduced in ~\cite{lin2017focal} over two classes (face and background). $ N_k^c $ is the number of anchors in $ P_k $ which participate in the classification loss computation. The regression loss $ L_r(t_i, t_i^*) $ is smooth L1 loss defined in ~\cite{girshick2015fast}. $ I(p_i^*=1) $ is the indicator function that limits the regression loss only focusing on the positively assigned anchors, and $ N_k^r=\sum_{i\in A_k}{I(p_i^*=1)} $. The attention loss $ L_a(m_k, m_k^*) $ is pixel-wise sigmoid cross entropy. $ m_k $ is the attention map generated per level, and $ m_k^* $ is the ground-truth described in Section~\ref{sec:attention func}.$ \lambda_1 $ and $ \lambda_2 $ are used to balance these three loss terms, here we simply set $ \lambda_1=\lambda_2=1 $.



\section{Experiments}
We use ResNet-50 as base model. All models are trained by SGD over 8 GPUs with a total of 32 images per mini-batch (4 images per GPU). Similar to~\cite{lin2017focal}, the four convolution layers attached to FPN are initialized with bias $ b=0 $ and Gaussian weight variance $ \sigma=0.01 $. For the final convolution layer of the classification subnet, we initiate them with bias $ b=?\log((1-\pi)/\pi) $ and $ \pi=0.01 $ here. Meanwhile, the initial learning rate is set as 3e-3. We sample 10k image patches per epoch. Models without data augmentation are trained for 30 epochs. With data augmentation, models are trained for 130 epochs, whose learning rate is dropped by 10 at 100 epochs and again at 120 epochs. Weight decay is 1e-5 and momentum is 0.9. Anchors with IoU $ \geq0.5 $ are assigned to positive class and anchors which have an IoU $ <0.4 $ with all ground-truth are assigned to the background class.

\subsection{Datasets}
The performance of FAN is evaluated across multiple face datasets: WiderFace and MAFA.

\textbf{WiderFace dataset~\cite{yang2016wider}:} WiderFace dataset contains 32, 203 images and 393, 703 annotated faces with a high degree of variability in scale, pose and occlusion. 158, 989 of these are chosen as train set, 39, 496 are in validation set and the rest are test set. The validation set and test set are split into 'easy', 'medium', 'hard' subsets, in terms of the difficulties of the detection. Due to the variability of scale, pose and occlusion, WiderFace dataset is one of the most challenge face datasets. Our FAN is trained only on the train set and evaluate on both validation set and test set. Ablation studies are performed on the validation set.

\textbf{MAFA dataset~\cite{Ge_2017_CVPR}:} MAFA dataset contains 30, 811 images with 35, 806 masked faces collected from Internet. It is a face detection benchmark for masked face, in which faces have vast various orientations and occlusion. Beside, this dataset is divided into masked face subset and unmasked face subset according to whether at least one part of each face is occluded by mask. We use both the whole dataset and occluded subset to evaluate our method.






\begin{table}[h]
\begin{center}
\begin{tabular}{|l|c|c|c|}
\hline
Method & easy & medium & hard \\
\hline
RetinaNet Setting & 92.6 & 91.2 & 63.4 \\
Dense Setting & 91.8 & 90.5 & 69.8 \\
Our FAN Baseline & 89.0 & 87.7 & 79.8 \\
\hline
\end{tabular}
\end{center}
\caption{The impact of anchor assign strategy on WiderFace validation set.}
\label{table: anchor setting}
\end{table}

\begin{figure*}[t]
\begin{center}
    \includegraphics[width=0.9\linewidth]{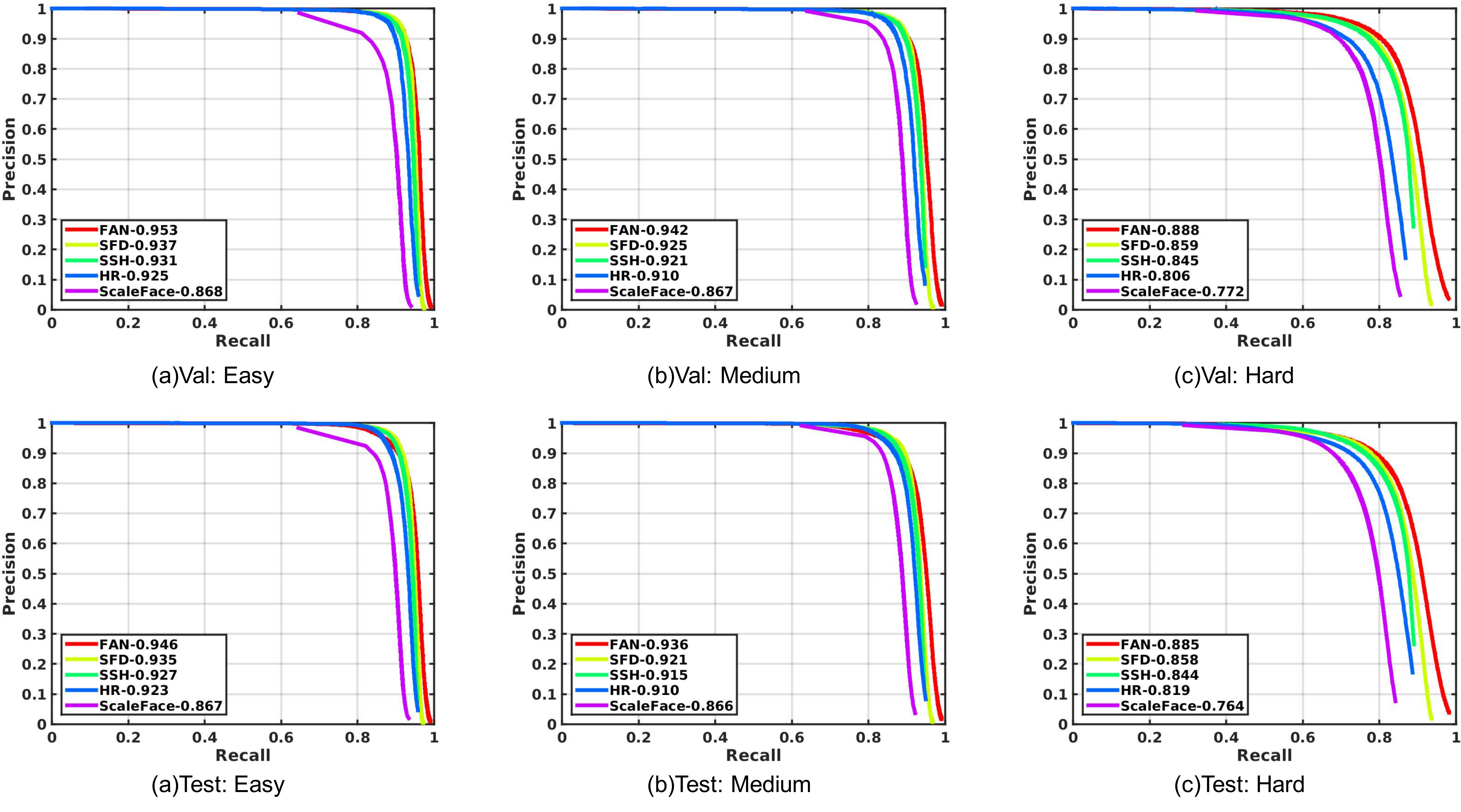}
\end{center}
    \caption{Precision-recall curves on WiderFace validation and test sets.} 
\label{fig:pr_curve}
\end{figure*}

\subsubsection{Anchor setting and assign}

We compare three anchor settings in Table~\ref{table: anchor setting}. For the \textbf{RetinaNet setting}, we follow the setting described in the paper~\cite{lin2017focal}. For \textbf{our FAN baseline}, we set our anchors from areas of $ 16^2 $ to $ 406^2 $ on pyramid levels. In addition, the aspect ratio is set to 1 and 1.5. Also, inspire of~\cite{zhang2017s3fd}, we choose an anchor assign rule with more cover rate. We uses 8 anchors per location spanning 4 scales (intervals are still fixed to $ 2^{1/3} $ so that areas from $ 8^2 $ to $ 645^2 $) and 2 ratios {1, 1.5}.  For the \textbf{dense setting}, it is the same as our FAN setting except we apply more dense scales from $ 8^2 $ to $ 645^2 $. Based on the results in Table~\ref{table: anchor setting}, we can see that anchor scale is important to detector performance and we can see that our setting is obviously superior to the setting in~\cite{lin2017focal}. Compared with the dense setting, we can see that anchor cover rate is not equal to the final detection performance as it may introduce a lot of negative windows due to the dense sampling.

\subsubsection{Attention mechanism}
As discussed in Section~\ref{sec:attention func}, we apply anchor-level attention mechanism to enhance the facial parts. We compare our FAN baseline with and without attention in Table~\ref{table: attention_wider} for the WiderFace val dataset. For the MAFA dataset?the results can be found in Table~\ref{table: attention_mafa}. Based on the experimental results, we can find that our attention can improve 1.1\% in WiderFace hard subset and 2\% in MAFA masked subset.

\begin{table}[htbp]
\begin{center}
\begin{tabular}{|l|c|c|c|}
\hline
Method & easy & medium & hard \\
\hline
baseline & 89.0 & 87.7 & 79.8 \\
attention & 88.4 & 88.3 & 80.9 \\
\hline
\end{tabular}
\end{center}
\caption{Result of attention mechanism comparison experiments on the WiderFace test set.}
\label{table: attention_wider}
\end{table}

\begin{table}[htbp]
\begin{center}
\begin{tabular}{|l|c|c|c|}
\hline
Method & occluded & all \\
\hline
baseline & 75.5 & 87.0 \\
attention & 76.5 & 88.3 \\
\hline
\end{tabular}
\end{center}
\caption{Result of attention mechanism comparison experiments on the MAFA test set.}
\label{table: attention_mafa}
\end{table}

\begin{table*}[htbp]
\begin{center}
\resizebox{\linewidth}{!}{
\begin{tabular}{c|ccccc|ccc}
\hline\hline
BaseNet & Dense anchor & Anchor assign & Attention & Data augmentation & Multi-scale & AP (easy) & AP (medium) & AP (hard) \\
\hline\hline
RetinaNet & & & & & & 92.6 & 91.2 & 63.4 \\
\hline
FAN & \checkmark & & & & & 91.8 & 90.5 & 69.8 \\
FAN & & \checkmark & & & & 89.0 & 87.7 & 79.8 \\
FAN & & \checkmark & \checkmark & & & 88.4 & 88.4 & 80.9 \\
FAN & & \checkmark & \checkmark & & \checkmark & 91.7 & 90.4 & 84.2 \\
FAN & & \checkmark & \checkmark & \checkmark & & 94.0 & 93.0 & 86.8 \\
FAN & & \checkmark & \checkmark & \checkmark & \checkmark & 95.3 & 94.2 & 88.8 \\
\hline
\end{tabular}
}
\end{center}
\caption{The ablation study of FAN on the WiderFace validation set.}
\label{tab:tech_path}
\end{table*}

\subsubsection{Data augmentation} 

According to the statistics from the WiderFace dataset, there are around 26\% of faces with occlusion. Among them, around 16\% is of serious occlusion. As we are targeting to solve the occluded faces, the number of training samples with occlusion may not be sufficient. Thus, we employ the the random crop data augmentation as discussed in Section~\ref{sec:data_aug}. The results can be found from Table~\ref{tab:tech_path}. The performance improvement is significant. Besides from the benefits for the occluded face, our random crop augmentation potentially improve the performance of small faces as more small faces will be enlarged after augmentation.

\subsubsection{WiderFace Dataset}

\begin{figure*}[tp]
\begin{center}
    \includegraphics[width=0.9\linewidth]{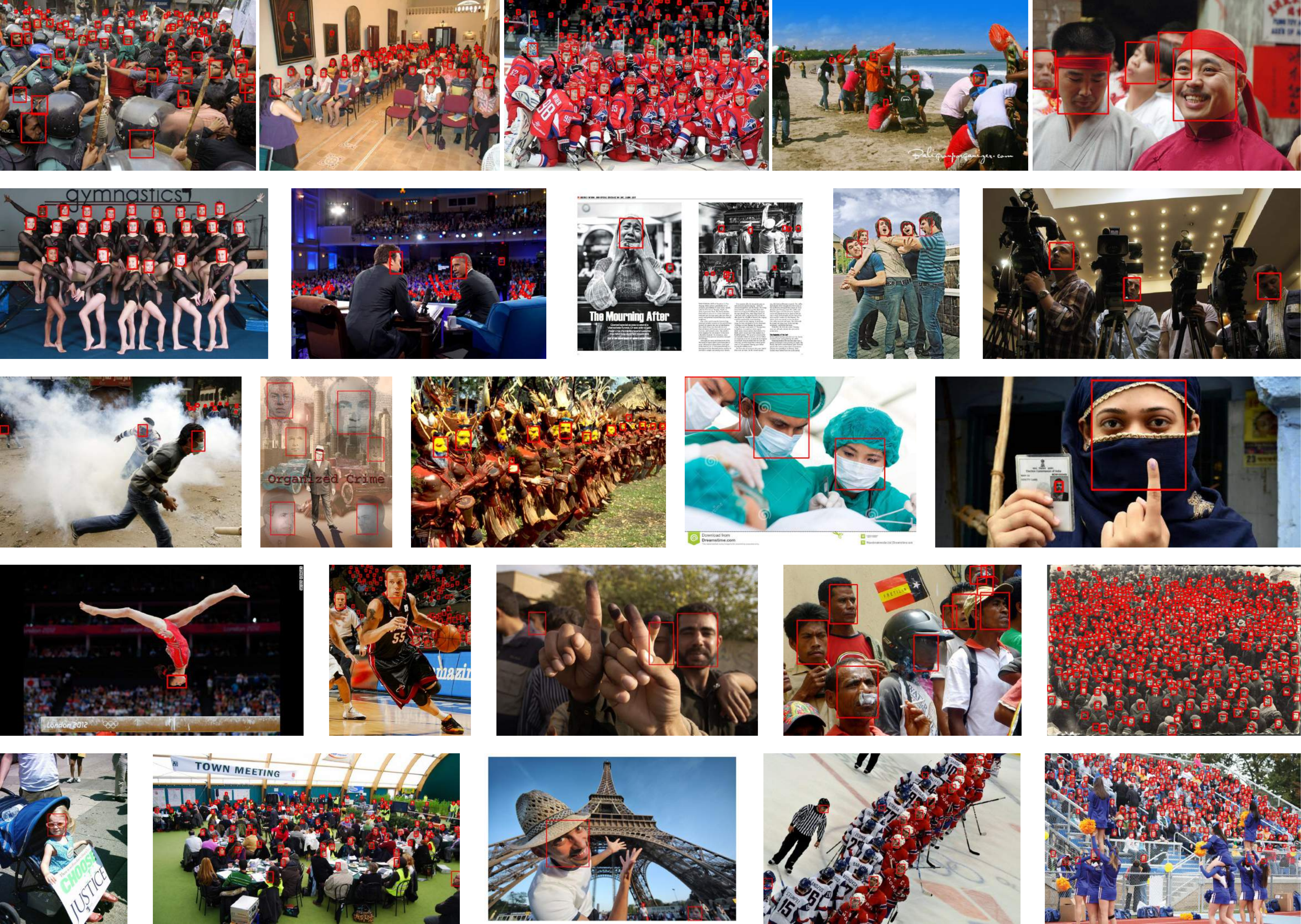}
\end{center} 
    \caption{Qualitative result of FAN on the validation set of the WiderFace dataset.}
\label{fig:widerface_vis}
\end{figure*}

We compare our FAN with the state-of-art detectors like SFD~\cite{zhang2017s3fd}, SSH~\cite{najibi2017ssh}, HR~\cite{Hu_2017_CVPR} and ScaleFace~\cite{yang2017face}. Our FAN is trained on WiderFace train set with data augmentation and tested on both validation and test set with multi-scale {600, 800, 1000, 1200, 1400}. The precision-recall curves and AP is shown in Figure~\ref{fig:pr_curve} and Table~\ref{table: widerface_result}. Our algorithm obtains the best result in all subsets, i.e. 0.953 (Easy), 0.942 (Medium) and 0.888 (Hard) for validation set, and 0.946 (Easy), 0.936 (Medium) and 0.885 (Hard) for test set. Considering the hard subset which contains a lot of occluded faces, we have larger margin compared with the previous state-art-results, which validates the effectiveness of our algorithm for the occluded faces. Example results from our FAN can be found in Figure~\ref{fig:widerface_vis}.


\begin{table}[t]
\begin{center}
\resizebox{\linewidth}{!}{
\begin{tabular}{|l|c|c|c|}
\hline
Method & AP(easy) & AP (medium) & AP(hard) \\
\hline
ACF~\cite{yang2014aggregate} & 69.5 & 58.8 & 29.0 \\
Faceness~\cite{yang2015facial} & 71.6 & 60.4 & 31.5 \\
LDCF+~\cite{ohn2016boost} & 79.7 & 77.2 & 56.4 \\
MT-CNN~\cite{zhang2016joint} & 85.1 & 82.0 & 60.7 \\
CMS-RCNN~\cite{zhu2017cms} & 90.2 & 87.4 & 64.3 \\
ScaleFaces~\cite{yang2017face} & 86.7 & 86.6 & 76.4 \\
HR~\cite{Hu_2017_CVPR} & 92.3 & 91.0 & 81.9 \\
SSH~\cite{najibi2017ssh} & 92.7 & 91.5 & 84.4 \\
SFD~\cite{zhang2017s3fd} & 93.5 & 92.1 & 85.8 \\
\hline
FAN & \textbf{94.6} & \textbf{93.6} & \textbf{88.5} \\
\hline
\end{tabular}
}
\end{center}
\caption{Comparison of FAN with state-of-art detectors on the test set of the WiderFace dataset.}
\label{table: widerface_result}
\end{table}

\subsubsection{MAFA Dataset} 
\begin{figure*}[bp]
\begin{center}
    \includegraphics[width=0.9\linewidth]{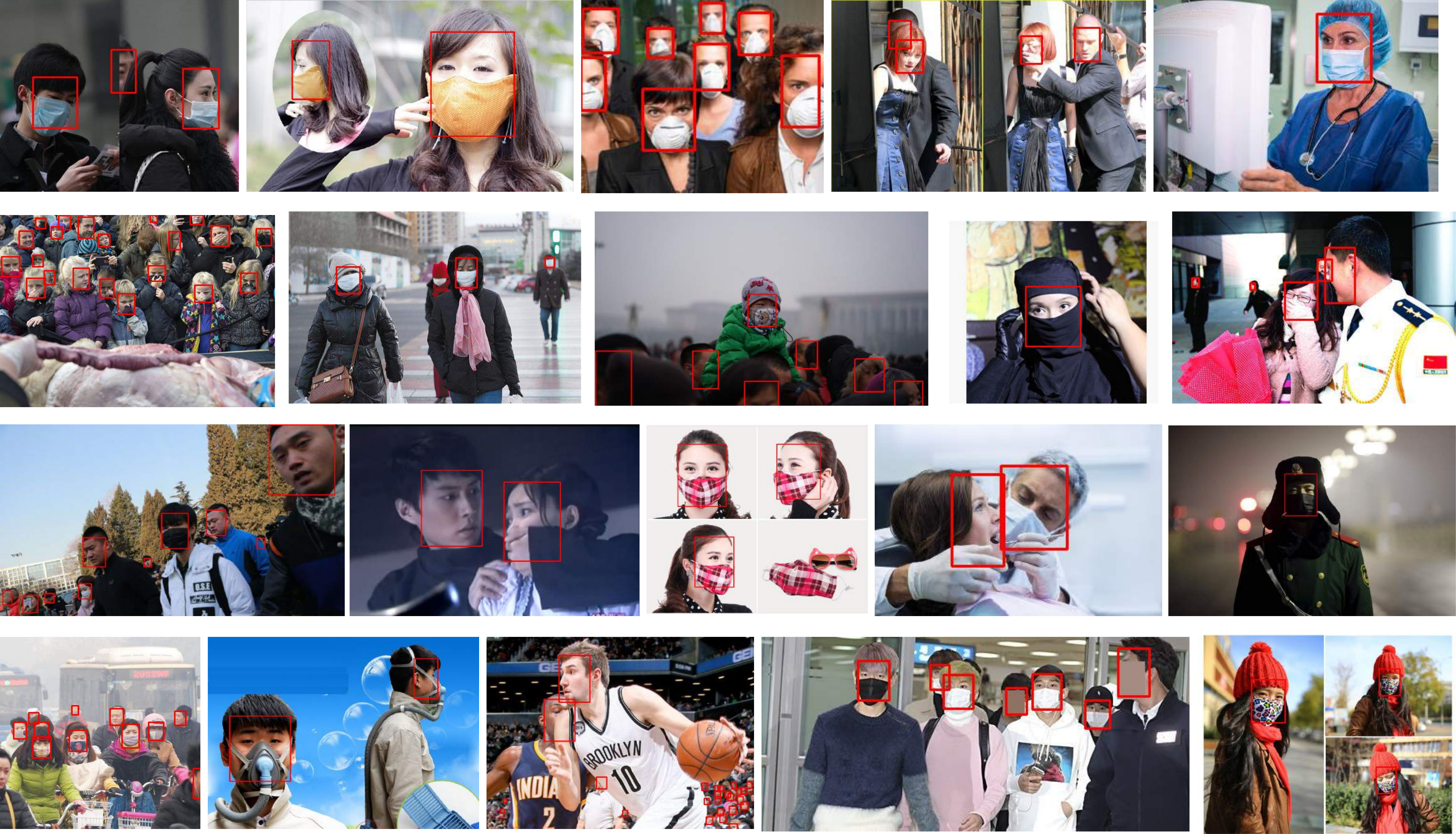}
\end{center} 
    \caption{Qualitative result of FAN on the validation set of the MAFA dataset.}
\label{fig:mafa_vis}
\end{figure*}

As MAFA dataset~\cite{Ge_2017_CVPR} is specifically designed for the occluded face detection, it is adopted to evaluate our algorithm. We compare our FAN with LLE-CNNs~\cite{Ge_2017_CVPR} and AOFD~\cite{chen2017masquer}. Our FAN is trained on WiderFace train set with data augmentation and tested on MAFA test set with scale {400, 600, 800, 1000}. The results based on average precision is shown in Table~\ref{table: mafa_result}. FAN significantly outperforms state-of-art detectors on MAFA test set with standard testing (IoU threshold = 0.5), which shows the promising performance on occluded faces. Example results from our FAN can be found in Figure~\ref{fig:mafa_vis}.

\begin{table}[t]
\begin{center}
\begin{tabular}{|l|c|c|}
\hline
Method & mAP \\
\hline
LLE-CNNs~\cite{Ge_2017_CVPR} & 76.4 \\
AOFD~\cite{chen2017masquer} & 77.3 \\
\hline
FAN & \textbf{88.3} \\
\hline
\end{tabular}
\end{center}
\caption{Comparison of FAN with state-of-art detectors on the test set of the MAFA dataset.}
\label{table: mafa_result}
\end{table}

\subsection{Inference Time}
Despite great performance obtained by our FAN, the speed of our algorithm is not compromised. As shown in Table~\ref{table: time}, our FAN detector can not only obtain the state-of-art results but also possess efficient computational speed. The computational cost is tested on a \emph{NIVIDIA TITAN Xp}. The min size $m$ means the shortest side of the images which are resized to $m$ by keeping the aspect ratio. Compared with the baseline results in Table~\ref{table: widerface_result}, when testing with short-side 1000, our FAN already outperforms state-of-art detectors like~\cite{najibi2017ssh}, \cite{zhang2017s3fd} and \cite{Hu_2017_CVPR}. 

\begin{table}[t]
\begin{center}
\resizebox{\linewidth}{!}{
\begin{tabular}{|c|c|c|c|c|c|}
\hline
Min size & 400 & 600 & 800 & 1000 & 1200\\
\hline
Time & 23.8ms & 36.1ms & 51.5ms & 67.9ms & 92.8ms \\
AP (hard) & 54.6 & 73.5 & 83.4 & 86.8 & 87.6 \\
\hline
\end{tabular}
}
\end{center}
\caption{The inference time and precision with respect to different input sizes for our FAN.}
\label{table: time}
\end{table}

\section{Conclusion}
In this paper, we are targeting the problem of face detection with occluded faces. We propose FAN detector which can integrate our specifically designed single-stage base net and our anchor-level attention algorithm. Based on our anchor-level attention, we can highlight the features from the facial regions and successfully relieving the risk from the false positives. Experimental results on challenging benchmarks like WiderFace and MAFA validate the effectiveness and efficiency of our proposed algorithm.

\clearpage

\clearpage

{\small
\bibliographystyle{ieee}
\bibliography{egbib}
}

\end{document}